\documentclass[conference]{IEEEtran}
\IEEEoverridecommandlockouts
\usepackage{cite}
\usepackage{amsmath,amssymb,amsfonts}
\usepackage{graphicx}
\usepackage{textcomp}
\usepackage{xcolor}
\usepackage{booktabs}
\usepackage{multirow}
\usepackage{colortbl}
\usepackage{array}
\usepackage{url}
\definecolor{veryhigh}{RGB}{76,153,0}
\definecolor{high}{RGB}{144,190,109}
\definecolor{medium}{RGB}{255,255,153}
\definecolor{low}{RGB}{253,174,97}
\definecolor{verylow}{RGB}{215,48,39}
\definecolor{zero}{RGB}{165,15,21}

\begin{document}
\title{Health-Conditioned Vision-Language-Action Models\\for Malfunction-Aware Robot Control}

\author{
\IEEEauthorblockN{Hüseyin ARSLAN\IEEEauthorrefmark{1}, Özgür ERKENT\IEEEauthorrefmark{1}}
\IEEEauthorblockA{\IEEEauthorrefmark{1}Computer Engineering, Hacettepe University, Ankara, Turkey \\
huseyinarslan25@hacettepe.edu.tr}

}

\maketitle

\begin{abstract}
Research on Vision Language Action (VLA) models has been increasing rapidly in recent years. Although some of them focus on detecting, preventing, and recovering from task failures, they usually don't deal with adapting to robot's physical failures. In real-life scenarios, most robots face physical degradations in various ways such as joint degradation, actuator failure, or weak gripper. We introduce malfunction-aware (health-conditioned) VLA that takes a health vector as an input that gives information about robots' joints' operation angle and torque capability, and adapts its predictions to complete the tasks with the degraded joints. To achieve this, we inject a Health Projector module to the VLA-Adapter architecture and train it on malfunction robot data we collected on the LIBERO environment \cite{liu2024libero}. We collect 128 teleoperated episodes on Libero-Spatial tasks. Our results show that, with a very lightweight addition, the model can learn to operate successfully with different configurations of degraded joints which the default pretrained VLA-Adapter's Libero-Spatial-Pro model cannot. The code and dataset will be available soon at \url{https://github.com/h-arslan/health-aware-vla}
\end{abstract}


\section{Introduction}

Vision-Language-Action (VLA) models have emerged as a highly promising topic for generalist robot control, combining Vision-Language Models (VLMs) with end-to-end action prediction architectures. Studies such as $\pi_0$ \cite{black2024pi0}, Project GR00T \cite{nvidia2024groot}, and VLA-Adapter \cite{chen2025vlaadapter} have demonstrated significantly high accuracy across various simulation and real-world tasks. They have shown the ability to operate in cross-embodiment setups and diverse scenarios, despite some existing limitations. Furthermore, with certain modifications, VLAs can run on edge devices that have limited computational resources and inference capabilities.

Along with these improvements, all the VLA approaches assume a robot hardware or simulation that is in perfect physical conditions. In real world, robots facing various deficiencies such as sensor drift, partially failures of actuators, lose torque of motors \cite{cully2015}. These kinds of situations will lead to task failure inevitably, because models, that are trained on healthy demonstrations of robot bodies, cannot adapt to new conditions.

This is a fundamental gap among the VLA studies, because unlike the classical controllers \cite{hogan1985impedance}, VLAs predict actions from learned representations of healthy robot embodiments. Some studies detect and recover from task failure semantically \cite{duan2024aha, safe2025, grisset2026ifailsense}, but no mechanism exists to adapt to joint or any other physical degradation among VLAs.

We address this gap by presenting malfunction aware VLA, an approach that combines pretrained VLA with an explicit health vector per joint $\mathbf{h} \in [0, 1]^J$ where $J$ is the number of joints and $h_j$ represents the health status, $h_j = 1.0$ means full health and $0.0$ indicates joint is fully locked. Proposed Health Projector maps this health vector into the model's latent space in order to enable action head adapts its predictions to physically degraded joints to complete the tasks successfully. Our main contributions can be listed as follows:
\begin{itemize}
\item We integrate a \textbf{health vector} into a pretrained VLA model with a small number of parameters (<1M) which result in a lightweight Health Projector improving model's adaptation capability to restricted joints significantly, and formulated the malfunction aware VLA problem.
\item \textbf{We introduce a data collection framework} that enables teleoperating LIBERO simulation tasks with mouse and keyboard by defining joint degradation with different levels.
\item \textbf{We present evaluation metrics that systematically measures the capability of VLA-Adapter} under conditions of different joint-degradation combinations on LIBERO Spatial tasks.
\end{itemize}

\section{Related Work}

\textbf{VLA models} can be defined as an extended version of VLMs that have ability of action predictions. RT-2 \cite{brohan2023rt2} is the first well-known study that presents the ability of action predictions with VLMs. OpenVLA \cite{kim2024openvla} released a 7B sized end-to-end model trained on OpenX Embodiment dataset \cite{team2024octo} that takes image and language input and predicts actions. The modifications on the backbones of VLA models, such as using different vision encoders and different VLM models \cite{yang2024qwen25, zhai2023siglip}, improved the performance in VLA's comes. Further performance improvement has been achieved with the usage of action prediction, data curation and training strategies. In OpenVLA \cite{kim2024openvla}, the model predicts actions in an autoregressive manner. $\pi_0$ \cite{black2024pi0}, Project GR00T \cite{nvidia2024groot}, and some other models use diffusion transformers and flow matching for action prediction.

\textbf{Failure Aware Vision Language Action Models.} Studies on fault tolerance \cite{maciejewski1990faulttolerant, zhang2013faulttolerant} in VLAs mainly focus on task based semantic failures and implicit approaches that focus on domain randomization \cite{tobin2017domainrandomization}. In a recent study, it is stated that, adaptation to cross embodiment and different physical failures is under a scaling law \cite{ai2025towards}. Our approach brings the idea of health projected action prediction with an explicit health descriptor to adapt to physical degradations, without requirement of multiple trainings of various scenarios for scaling.

\section{Approach}

\subsection{Problem Formulation}

As we used LIBERO Spatial task suite, we consider a 7-DOF manipulator that can fail anytime by joint degradation. Each joint $j$ is described by a health value $h_j \in [0, 1]$, where $h_j = 1$ means full health and $h_j = 0$ indicates a completely locked joint. Intermediate values reduce both the actuator gain and the reachable angle range of joint by the factor $(1 - w)$, with $w = 1 - h_j$. The complete robot state is therefore enhanced with a \emph{health vector} $\mathbf{h} = [h_0, h_1, \dots, h_6] \in [0,1]^7$.

A standard VLA policy $\pi_\theta$ maps visual observations $\mathbf{o}_t$ and a language instruction $l$ to a chunk of $C$ future actions:
\begin{equation}
\hat{\mathbf{a}}_{t:t+C} = \pi_\theta(\mathbf{o}_t, l)
\end{equation}
Since this policy is trained on episodes from a healthy robot, it cannot adapt its action predictions for degraded joints and mobility failures.

We propose a new policy that is \emph{health-conditioned}:
\begin{equation}
\hat{\mathbf{a}}_{t:t+C} = \pi_\theta(\mathbf{o}_t, l, \mathbf{h})
\end{equation}
where the health vector $\mathbf{h}$ is projected into the model's latent space  via an MLP and fused with the action prediction head, allowing the model to adapt its motion plan to the current joint capabilities.

\subsection{Architecture}

We use \textbf{VLA-Adapter Pro} \cite{chen2025vlaadapter} as base model which pairs a frozen Qwen2.5-0.5B language model with a fused DINOv2--SigLIP vision encoder and an action head trained with L1 regression. We add a lightweight module and modify the action head's attention to explicitly integrate health information of robot.

\textbf{Base VLA-Adapter Pro.}
The action head is a 24-block residual MLP network. Each block contains an 8-head attention layer that attends over three sources: (i)~\emph{self} tokens (the evolving action latent), (ii)~\emph{adapter} tokens (learnable action queries concatenated with proprioceptive features), and (iii)~\emph{task} tokens (vision--language features from the VLM backbone). Separate key--value projections are used per source with rotary position embeddings (RoPE), and a learned gating controls task attention strength. The head predicts a chunk of $C{=}8$ future 7-DoF end-effector delta actions.

\textbf{Health Projector.}
We proposed a two-layer MLP that maps the 7D health vector into the backbone LLM embedding space ($\mathbb{R}^{896}$):
\begin{equation}
\mathbf{f}_h = W_2 \, \text{GELU}(W_1 \mathbf{h} + b_1) + b_2
\end{equation}
The final layer ($W_2, b_2$) is \emph{zero-initialized} so that $\mathbf{f}_h = \mathbf{0}$ at the start of training, pretrained action head is not be affected. The health features are added element-wise to the proprioceptive token before it enters the attention blocks, so the adapter tokens carry both information of proprioceptive and health without changing the token count. This additive design means the pretrained attention pattern is preserved at initialization: $\mathbf{p} + \mathbf{0} = \mathbf{p}$.

\textbf{Parameter budget.}
The Health Projector adds 810K parameters and Action Queries adds 57K parameters. These sum up to only 900K trainable parameters which is easy to finetune with limited resources and does not increase inference budget.

\begin{figure}[t]
\centering
\includegraphics[width=\linewidth]{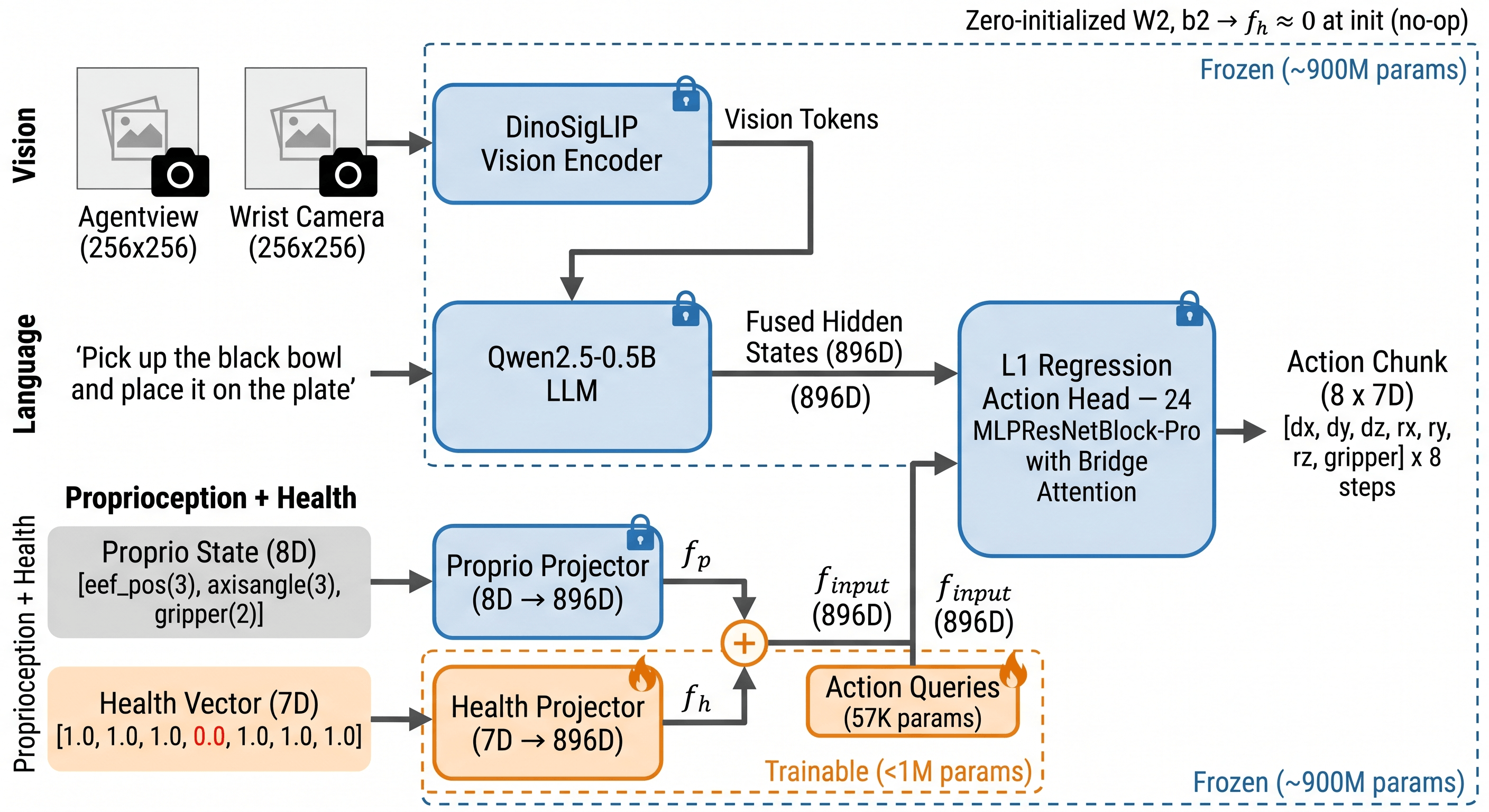}
\caption{Health-conditioned VLA architecture. The Health Projector (orange) maps the 7D health vector into the model's latent space and adds it to the proprioceptive features. Only the Health Projector and action queries are trained; all other components remain frozen.}
\label{fig:architecture}
\vspace{-3mm}
\end{figure}

\subsection{Data Collection}

As our base model VLA-Adapter has pretrained weights on different LIBERO Benchmarks, we created a dataset on LIBERO Spatial tasks. We created an interface to operate in LIBERO Simulation using OSC (Open Space Control) by controlling robot body with mouse and wrist, hand, gripper with keyboard. Joint malfunctions are simulated by constraining actuator gain and operation angle scaled by $1 - w$, where $w = 0$ is healthy and $w = 1$ means joint is fully locked. We evaluated the VLA-Adapter on 6 joints with different malfunction levels; then collected teleoperated data on conditions that VLA-Adapter fails. Malfunction levels are: 0.3, 0.5, 0.7 and 0.9. On each level configuration, we collected 2 episodes. To preserve data consistency, we added episodes from VLA-Adapter's non-degraded runs with full healthy vector and degraded runs where VLA-Adapter was able to succeed. Each episode records: dual-camera RGB observations (256$\times$256), 7D end-effector delta actions, 8D proprioceptive state (end-effector pose + gripper), and the 7D health vector. In total, we collected 128 malfunction episodes on different joint-malfunction configurations, and 50 healthy episodes using pretrained VLA-Adapter Libero Spatial Pro model.

\subsection{Training Setup}

We used a single NVIDIA L40S GPU to finetune our health conditioned VLA architecture. As we use VLA-Adapter-Pro as base architecture, we keep VLM backbone, vision encoder and Bridge Attention Adapter frozen. Only Health Projector and Action Queries are trainable which sums up to 900K trainable params,  from 57K action queries and 800K Health Projector. We used AdamW optimizer with learning rate 2e-4. Training completed at 15K step with batch size 8 and gradient accumulator over 2 steps, gradient clipping at norm 1.0. Actions are normalized to [-1,1] using BOUNDS Q99 statistics and converted to the RLDS convention to match the pretrained action head’s output space. Learning rate was warming up linearly over with step number 500 and decays by 10x at step 10K. Different image augmentations are used to increase the data robustness across the tasks and objects. L1 regression is used for 8 action chunk loss.

\section{Experiments}

\subsection{Evaluation Protocol}

For evaluation and finetuning, we only used LIBERO Spatial tasks, each tasks requires picking up the black bowl from different places and putting it on the plate. It measures the spatial understanding of the robot. Since semantic requirements of the tasks is not relevant to our study, we pick the LIBERO Spatial tasks. For baseline, we ran the pretrained VLA-Adapter without health projector on every task x joint x malfunction combination (7 joints × 4 weakness levels {0.3, 0.5, 0.7, 0.9}), including completely healthy setup. Due to runtime constraints, we run 4 episodes each. For our proposed health-conditioned model, we run 10 episodes for each task x joint x condition pair. Our primary metric is binary task success rate similar to VLA-Adapter.

\subsection{Baseline Vulnerability Analysis}

Table~\ref{tab:baseline} shows the pretrained VLA-Adapter LIBERO-Spatial-Pro model evaluated on every single-joint degradation. The baseline achieves 97.5\% healthy success rate, which confirms strong nominal performance. However, joint degradation exposes sharp vulnerabilities. J1 (Shoulder) drops to 45\% at $w{=}0.3$ and falls to 0\% for $w \geq 0.7$; J3 (Elbow) follows a similar pattern at 42.5\% ($w{=}0.3$) to 0\% ($w \geq 0.5$). These two proximal joints bear the largest kinematic load for the pick-and-place motion, and even slight restrictions produce trajectories that the pretrained policy cannot compensate for. Some other joints (J4--J6) are more resilient: J4 preserves 97.5\% at all weakness levels, while J5 and J6 stay above 65\% even at $w{=}0.9$. J0 (Base) and J2 (Upperarm) remain largely unaffected (99.4\% and 88.1\% average). These results motivate health-conditioned finetuning---the baseline has no mechanism to adapt to degraded kinematics.

\begin{table}[t]
\caption{VLA-Adapter Baseline: Success Rate (\%) Under Joint Malfunctions (No Health Conditioning)}
\label{tab:baseline}
\centering
\begin{tabular}{l|cccc|c}
\toprule
\textbf{Joint} & \textbf{w=0.3} & \textbf{w=0.5} & \textbf{w=0.7} & \textbf{w=0.9} & \textbf{Avg.} \\
\midrule
J0 (Base)& 100.0 & 100.0 & 97.5 & 100.0 & 99.4 \\
J1 (Shoulder) & 45.0& 15.0& 0.0& 0.0 & 15.0 \\
J2 (Upperarm) & 95.0& 97.5& 87.5 & 72.5& 88.1 \\
J3 (Elbow)& 42.5& 0.0 & 0.0& 0.0 & 10.6 \\
J4 (Forearm)& 97.5 & 97.5& 97.5 & 97.5& 97.5 \\
J5 (Wrist)& 100.0 & 95.0& 65.0 & 67.5& 81.9 \\
J6 (Hand)& 100.0 & 97.5& 90.0 & 72.5& 90.0 \\
\midrule
\multicolumn{5}{l|}{\textit{Healthy baseline}} & 97.5 \\
\bottomrule
\end{tabular}
\vspace{-2mm}
\end{table}

\subsection{Health-Conditioned Results}

Table~\ref{tab:ours} compares the baseline pretrained model and our health-conditioned model across all joints and weakness levels on all tasks. Our model achieves 99.0\% healthy success rate, similar to the baseline (97.5\%), which confirms that the zero-initialized Health Projector preserves nominal performance when $\mathbf{h}=\mathbf{1}$.

On the critical joints, where the baseline collapses, our model shows considerable improvement. J1 (Shoulder) increases from 45\%$\rightarrow$89\% at $w{=}0.3$ and from 15\%$\rightarrow$22\% at $w{=}0.5$, and recovers a non-zero 2\% at $w{=}0.7$ where the baseline scores 0\%. J3 (Elbow) improves from 0\%$\rightarrow$34\% at $w{=}0.5$ and gains 3\% at $w{=}0.7$. For joints that were already succeeding, our model maintains competitive performance: J2 improves from 72.5\%$\rightarrow$83\% at $w{=}0.9$, and J5 improves across all high-weakness levels (65\%$\rightarrow$82\% at $w{=}0.7$, 67.5\%$\rightarrow$79\% at $w{=}0.9$). J6 shows a notable shift from 72.5\%$\rightarrow$82\% at $w{=}0.9$.

We observed a trade-off on joints that the baseline is already working well. J0 drops from near-perfect to 71--91\% and J4 from 97.5\% to 89--97\%. This could be solved by adding data saved using pretrained models action predictions into the dataset. Since our dataset contains healthy rollouts of pretrained model and teleoperated episodes where VLA-Adapter fails, it is expected that our model may have inconsistencies at some combinations due to lack of pretrained episodes.

The per-task breakdown in Table~\ref{tab:pertask} reveals where gains concentrate. On tasks where baseline completely fails (e.g., T5 J1 $w{=}0.3$: 0\%$\rightarrow$30\%; T8 J3 $w{=}0.5$: 0\%$\rightarrow$80\%), our model learns compensatory strategies. Conversely, some task--joint combinations remain at 0\% for both models (J1 and J3 at $w{=}0.9$), indicating physical limits that no action adaptation can overcome.

\begin{table}[t]
\caption{Success Rate (\%) Comparison: Baseline vs. Health-Conditioned Model}
\label{tab:ours}
\centering
\begin{tabular}{l|l|cccc}
\toprule
\textbf{Joint} & \textbf{Model} & \textbf{w=0.3} & \textbf{w=0.5} & \textbf{w=0.7} & \textbf{w=0.9} \\
\midrule
\multirow{2}{*}{J0 (Base)}
& Baseline & 100.0 & 100.0 & 97.5& 100.0 \\
& Ours & 91.0& 86.0& 72.0& 71.0\\
\midrule
\multirow{2}{*}{J1 (Shoulder)}
& Baseline & 45.0& 15.0& 0.0 & 0.0\\
& Ours & \textbf{89.0} & \textbf{22.0} & \textbf{2.0} & 0.0 \\
\midrule
\multirow{2}{*}{J2 (Upperarm)}
& Baseline & 95.0& 97.5& 87.5& 72.5\\
& Ours& \textbf{96.0} & 94.0& 85.0& \textbf{83.0} \\
\midrule
\multirow{2}{*}{J3 (Elbow)}
& Baseline & 42.5& 0.0& 0.0& 0.0 \\
& Ours& 39.0& \textbf{34.0} & \textbf{3.0} & 0.0 \\
\midrule
\multirow{2}{*}{J4 (Forearm)}
& Baseline & 97.5& 97.5& 97.5& 97.5\\
& Ours& 97.0& 90.0& 95.0& 89.0\\
\midrule
\multirow{2}{*}{J5 (Wrist)}
& Baseline & 100.0 & 95.0& 65.0& 67.5\\
& Ours & 90.0& 86.0& \textbf{82.0} & \textbf{79.0} \\
\midrule
\multirow{2}{*}{J6 (Hand)}
& Baseline & 100.0 & 97.5& 90.0& 72.5\\
& Ours & 71.0& 78.0& 79.0& \textbf{82.0} \\
\midrule
\multicolumn{2}{l|}{\textit{Healthy baseline}} & \multicolumn{4}{c}{Baseline: 97.5\% \quad Ours: 99.0\%} \\
\bottomrule
\end{tabular}
\vspace{-2mm}
\end{table}

\begin{table*}[t]
\caption{Per-Task Success Rate (\%) Under All Joint Malfunctions on LIBERO-Spatial (T0--T9). B = Baseline, O = Ours. J1 $w$=0.9 and J3 $w$=0.9 omitted (both models 0\% on all tasks). Bold indicates our model outperforms the baseline.}
\label{tab:pertask}
\centering
\scriptsize
\setlength{\tabcolsep}{2pt}
\begin{tabular}{c|c|c|cccc|ccc|cccc|ccc|cccc|cccc|cccc}
\toprule
& & & \multicolumn{4}{c|}{J0 (Base)} & \multicolumn{3}{c|}{J1 (Shldr)} & \multicolumn{4}{c|}{J2 (Uparm)} & \multicolumn{3}{c|}{J3 (Elbow)} & \multicolumn{4}{c|}{J4 (Forearm)} & \multicolumn{4}{c|}{J5 (Wrist)} & \multicolumn{4}{c}{J6 (Hand)} \\
\textbf{T} & & \textbf{H} & .3 & .5 & .7 & .9 & .3 & .5 & .7 & .3 & .5 & .7 & .9 & .3 & .5 & .7 & .3 & .5 & .7 & .9 & .3 & .5 & .7 & .9 & .3 & .5 & .7 & .9 \\
\midrule
\multirow{2}{*}{0}
& B & 100 & 100 & 100 & 100 & 100 & 25 & 50 & 0 & 100 & 100 & 100 & 100 & 100 & 0 & 0 & 100 & 100 & 100 & 100 & 100 & 100 & 75 & 50 & 100 & 100 & 100 & 100 \\
& O & 100 & 100 & 100 & 100 & 100 & \textbf{90} & 30 & 0 & 100 & 90 & 100 & 100 & 90 & \textbf{70} & 0 & 100 & 100 & 100 & 100 & 100 & 100 & \textbf{90} & \textbf{100} & 100 & 100 & 100 & 100 \\
\midrule
\multirow{2}{*}{1}
& B & 100 & 100 & 100 & 100 & 100 & 0 & 25 & 0 & 100 & 100 & 100 & 50 & 50 & 0 & 0 & 100 & 100 & 100 & 75 & 100 & 75 & 50 & 100 & 100 & 100 & 50 & 25 \\
& O & 100 & 80 & 70 & 20 & 40 & \textbf{100} & \textbf{60} & 0 & 80 & 90 & 60 & \textbf{70} & 50 & \textbf{30} & \textbf{20} & 80 & 60 & 80 & 50 & 90 & \textbf{100} & \textbf{100} & 100 & 10 & 50 & 30 & \textbf{60} \\
\midrule
\multirow{2}{*}{2}
& B & 100 & 100 & 100 & 100 & 100 & 0 & 25 & 0 & 100 & 100 & 100 & 100 & 25 & 0 & 0 & 100 & 100 & 100 & 100 & 100 & 100 & 25 & 25 & 100 & 100 & 100 & 100 \\
& O & 100 & 100 & 100 & 90 & 80 & \textbf{100} & \textbf{80} & 0 & 100 & 100 & 100 & 100 & \textbf{60} & \textbf{70} & 0 & 100 & 100 & 100 & 100 & 100 & 100 & \textbf{100} & \textbf{100} & 100 & 100 & 100 & 100 \\
\midrule
\multirow{2}{*}{3}
& B & 100 & 100 & 100 & 100 & 100 & 100 & 0 & 0 & 100 & 100 & 100 & 100 & 75 & 0 & 0 & 100 & 100 & 100 & 100 & 100 & 100 & 50 & 100 & 100 & 100 & 100 & 100 \\
& O & 100 & 100 & 100 & 100 & 100 & 90 & 0 & 0 & 100 & 100 & 100 & 100 & 30 & 0 & 0 & 100 & 100 & 100 & 100 & 100 & 80 & \textbf{60} & 70 & 90 & 80 & 100 & 100 \\
\midrule
\multirow{2}{*}{4}
& B & 100 & 100 & 100 & 100 & 100 & 75 & 25 & 0 & 100 & 100 & 100 & 75 & 0 & 0 & 0 & 100 & 100 & 100 & 100 & 100 & 100 & 100 & 100 & 100 & 100 & 50 & 0 \\
& O & 100 & 80 & 70 & 60 & 60 & \textbf{100} & 10 & 0 & 100 & 90 & 90 & \textbf{100} & \textbf{10} & \textbf{20} & 0 & 100 & 90 & 90 & 80 & 100 & 100 & 90 & 80 & 0 & 10 & 0 & \textbf{10} \\
\midrule
\multirow{2}{*}{5}
& B & 75 & 100 & 100 & 100 & 100 & 0 & 0 & 0 & 75 & 75 & 50 & 25 & 75 & 0 & 0 & 75 & 75 & 75 & 100 & 100 & 75 & 50 & 0 & 100 & 75 & 100 & 75 \\
& O & \textbf{100} & 100 & 90 & 90 & 100 & \textbf{30} & \textbf{20} & 0 & \textbf{100} & \textbf{100} & \textbf{100} & \textbf{90} & 70 & \textbf{70} & \textbf{10} & \textbf{100} & \textbf{90} & \textbf{100} & 90 & 70 & 50 & \textbf{60} & \textbf{30} & 100 & \textbf{100} & 100 & \textbf{100} \\
\midrule
\multirow{2}{*}{6}
& B & 100 & 100 & 100 & 75 & 100 & 75 & 0 & 0 & 100 & 100 & 100 & 100 & 25 & 0 & 0 & 100 & 100 & 100 & 100 & 100 & 100 & 100 & 100 & 100 & 100 & 100 & 100 \\
& O & 100 & 100 & 100 & \textbf{100} & 100 & \textbf{80} & 0 & 0 & 100 & 100 & 100 & 100 & 10 & 0 & 0 & 100 & 100 & 100 & 100 & 100 & 100 & 90 & 90 & 100 & 100 & 100 & 100 \\
\midrule
\multirow{2}{*}{7}
& B & 100 & 100 & 100 & 100 & 100 & 0 & 0 & 0 & 100 & 100 & 75 & 75 & 50 & 0 & 0 & 100 & 100 & 100 & 100 & 100 & 100 & 100 & 0 & 100 & 100 & 100 & 100 \\
& O & 100 & 60 & 60 & 30 & 10 & \textbf{100} & 0 & \textbf{20} & 100 & 90 & 60 & 30 & 0 & 0 & 0 & 100 & 80 & 90 & 80 & 50 & 40 & 40 & \textbf{30} & 100 & 100 & 100 & 100 \\
\midrule
\multirow{2}{*}{8}
& B & 100 & 100 & 100 & 100 & 100 & 75 & 0 & 0 & 75 & 100 & 50 & 0 & 25 & 0 & 0 & 100 & 100 & 100 & 100 & 100 & 100 & 0 & 100 & 100 & 100 & 100 & 25 \\
& O & 90 & 90 & 80 & 50 & 20 & \textbf{100} & \textbf{10} & 0 & \textbf{90} & 80 & \textbf{60} & \textbf{50} & \textbf{50} & \textbf{80} & 0 & 90 & 80 & 90 & 90 & 90 & 90 & \textbf{100} & 90 & 20 & 40 & 60 & \textbf{50} \\
\midrule
\multirow{2}{*}{9}
& B & 100 & 100 & 100 & 100 & 100 & 100 & 25 & 0 & 100 & 100 & 100 & 100 & 0 & 0 & 0 & 100 & 100 & 100 & 100 & 100 & 100 & 100 & 100 & 100 & 100 & 100 & 100 \\
& O & 100 & 100 & 90 & 80 & 100 & 100 & 10 & 0 & 90 & 100 & 80 & 90 & \textbf{20} & 0 & 0 & 100 & 100 & 100 & 100 & 100 & 100 & 90 & 100 & 90 & 100 & 100 & 100 \\
\bottomrule
\end{tabular}
\vspace{-2mm}
\end{table*}

\subsection{Analysis}

\textbf{Kinematic criticality determines difficulty.} As we presented in results, Joints 1 (Shoulder) and 3 (Elbow) with degradations are hard to solve because they control the main reaching and lifting motions. At high weakness levels, completing  a task is impossible due to drastic shrinkage of reachable workspace.

\textbf{Health conditioning enables adaptive strategies.} The Table 3 shows that on some tasks, our model achieves significantly large gains (T5 J2: 25→90 at w=0.9; T8 J3: 0→80 at w=0.5), suggesting that the model adapted task specific different trajectories that solves the degraded joint’s reaching problem. At some other tasks, improvement is limited because task does not have much motion alternatives to create.

\textbf{Cost of generalization.} As we collected our data by ourselves, dataset has highly limited coverage among the possibilities of malfunction level - joint combinations. This brings some results that model seemed to forget prior information and not able to adapt truly. This can be solved by creating a broader malfunction dataset and covering as much degradation combination as possible.

\textbf{Data efficiency.} We used only 178 episodes (128 malfunction + 50 healthy) to train the model and yet it achieves meaningful recovery on joints that pretrained model entirely fails on. Health Projector showed that explicit health conditioning is a highly parameter efficient approach for malfunction adaptation.

\section{Discussion and Future Work}

This work demonstrates that VLA models can be made aware of physical degradations through a simple, parameter-efficient health conditioning mechanism. By injecting an explicit health vector via a zero-initialized projector, we enable the pretrained VLA-Adapter to adapt its action predictions to degraded joints without sacrificing nominal performance.

\textbf{Limitations.} Our current approach is limited with only single joint degradation on only LIBERO task suite. Also it requires adding an explicit health vector. We believe that a module that able to generalize various degradations without any external health info can be developed.

\textbf{Multi joint degraded data.} Since our approach supports multi joint degradations, this work can be done but as we collect our data via teleoperation. It requires a data collection effort to build such a dataset.

\textbf{Implicit health estimation.} Our proposed architecture requires feeding health vector explicitly. A mechanism that learns from robot’s actions and reactions can be developed. That kind of work will bring zero shot adaptation to degradation and capability of being aware to malfunction knowledge.

\textbf{Scaling to diverse and real embodiments.} As our method is only finetuned and evaluated at simulated environments and robots, executing this approach on different real life robot embodiments remains as an open problem. 

\bibliographystyle{IEEEtran} 
\bibliography{egbib}

\end{document}